\documentclass[runningheads]{llncs}
\usepackage[english]{babel}
\usepackage{graphicx}
\usepackage{amssymb}
\usepackage{amsmath}
\usepackage[table]{xcolor}
\usepackage{multirow}
\usepackage{tabularx}
\usepackage{hyperref}
\usepackage{mymacros}
\usepackage{adjustbox}
\addto\extrasenglish{%
}

\begin{document}
%
% \title{Digital Twins of Organizations for Impact Analysis of Process-Aware Information Systems Updates}
% \title{Impact Analysis of Information Systems Updates Using Digital Twins of Organizations}
%
\title{Analyzing Process-Aware Information System Updates Using Digital Twins of Organizations\thanks{This work is supported by the Alexander von Humboldt (AvH) Stiftung and the 0000 Project Fund (Project n. 1.220047.01) of UNIST (Ulsan National Institute of Science \& Technology).}}

\titlerunning{Impact Analysis Using Digital Twins of Organizations}
\author{Gyunam Park\inst{1}\orcidID{0000-0001-9394-6513} 
\and
Marco Comuzzi\inst{2}\orcidID{0000-0002-6944-4705}
\and
Wil M. P. van der Aalst\inst{1}\orcidID{0000-0002-0955-6940}
}
% \author{Gyunam Park\inst{1} 
% \and
% Marco Comuzzi\inst{2}
% \and
% Wil M. P. van der Aalst\inst{1}
% }
%
\authorrunning{Park et al.}
\institute{Process and Data Science Group (PADS), RWTH Aachen University, Germany \\ \email{\{gnpark,wvdaalst\}@pads.rwth-aachen.de}
\and
Department of Industrial Engineering, UNIST, Korea \\ \email{mcomuzzi@unist.ac.kr}
}
\maketitle              % typeset the header of the contribution
\begin{abstract}
Digital transformation often entails small-scale changes to information systems supporting the execution of business processes.
These changes may increase the operational frictions in process execution, which decreases the process performance.
The contributions in the literature providing support to the tracking and impact analysis of small-scale changes are limited in scope and functionality.
In this paper, we use the recently developed Digital Twins of Organizations (DTOs) to assess the impact of (process-aware) information systems updates. 
More in detail, we model the updates using the configuration of DTOs and quantitatively assess different types of impacts of information system updates (structural, operational, and performance-related).
% Specifically, we model the updates using the configuration of DTOs, and we show how DTOs can be used to quantitatively assess different types of impacts of information systems updates (structural, operational and performance-related). 
We implemented a prototype of the proposed approach. 
Moreover, we discuss a case study involving a standard ERP procure-to-pay business process. 
\keywords{Business Process  \and System Update \and Impact \and Digital Twin.}
\end{abstract}

\section{Introduction}

Process-Aware Information Systems (PAIS), such as ERP and CRM, play a key role in modern organizations, underpinning the execution of business processes~\cite{dumas2005process}.
As the environment surrounding an organization dynamically changes and the competition becomes more intensive, a demand to accordingly change the implementation of PAIS may arise.
%ERP post-implementation deals with such demands by modifying system designs and functionalities of an ERP system~\cite{xxx}.
%While reshaping process behaviors in an organization, the aim of ERP post-implementation changes is to improve the overall performance of organizational operations. 

%ERP post-implementation often deals with large-scale changes, which can involve the (re-)configuration of entire ERP modules or business processes~\cite{xxx}. 
Modern digital transformation often involves frequent and small-scale changes, or \emph{updates}, to information systems, which are vital to enable continuous adaptation to dynamic business environments~\cite{mendling2020building}.
The need for such capabilities is illustrated by the measures needed to handle the Covid-19 pandemic.
For instance, organizations had to devise measures to adapt to varying degrees of home-working workforce, or to variations of sales and orders that deviate extensively and unexpectedly from the usual seasonality.

PAIS updates affect the operations of an organization.
However, due to the complex design and deep operational pervasiveness of modern PAIS, it is challenging to assess the impact of PAIS updates.
Contributions in the PAIS literature in this direction are limited.
Existing approaches mainly take a model-driven view on impact analysis~\cite{soffer2003erp,parhizkar2017impact,comuzzi2012measures}, and concrete implementations are missing.
Taking a design science standpoint~\cite{wieringa2014design}, we address this research gap by developing (software) artifacts supporting the impact analysis of PAIS updates.

In this work, we distinguish specifically among three types of impacts: \textit{structural}, \textit{operational}, and \emph{performance} impacts.
At the structural level, it is important to assess the magnitude of the scope of a proposed update, such as the number of business objects and business functions that it affects. 
%impact of a change  occur when changes in an ERP system do not conform to the system design or trigger side-effects. 
At the operational level, the impact of a PAIS update concerns the running instances of business processes involving the business objects and functions subject to change. For instance, removing the need to collect some customer information for privacy reasons may affect those customers for whom the information has been collected already.
%For instance, adjusting the minimum price of procurement orders from 10 EUR to 100 EUR to reduce the number of non-profitable orders may affect not only the incoming orders, but also payment policies, e.g., making a credit payment control policy for orders lower than 50 EUR obsolete. 
%Moreover, it may affect the orders lower than 100 Euros waiting for an order confirmation placed following the old business rule.
Most importantly, PAIS updates may impact on diagnostic measures of business processes: for instance, decreasing the level of tolerated mismatch between invoices and payments may slow down the processing of procurement cases, which in turn decreases the throughput of the procurement process.

To analyze the structural/operational/performance impact, we rely on Digital Twins of Organizations (DTOs).
A DTO is a digital replication of an organization's operations, providing a transparent view to the business processes of the organization and enabling process analysts to analyze existing operational frictions and identify improvement opportunities.
Moreover, DTOs allow to monitor business processes and trigger management actions, e.g., adding more resources and configuring business rules if improvements are available.

Specifically, we use a \textit{Digital-Twin Interface Model} (DT-IM)~\cite{park-digitaltwin} that (i) represents PAISs based on Object-Centric Petri Nets (OCPNs)~\cite{DBLP:journals/fuin/AalstB20} and (ii) models PAIS updates with the notion of \textit{actions}.
We compute the impact of a PAIS update with the DT-IM by analyzing the elements of the OCPN that it modifies (structural impacts), current states affected by the update (operational impacts), and the new value of diagnostic measures after the update (performance impacts).

From a design science standpoint, the practical relevance of the problem that we address lies with the increasing relevance of digital transformation for modern enterprises~\cite{mendling2020building}. 
The rigor of the design process is ensured by extending a sound conceptual model of DTOs already published in the literature, i.e., DT-IMs.
% Specifically, we use a \textit{Digital-Twin Interface Model} (DT-IM)~\cite{park-digitaltwin} that (i) represents PAISs based on Object-Centric Petri Nets (OCPNs)~\cite{DBLP:journals/fuin/AalstB20} and (ii) models PAIS updates with the notion of \textit{actions}.
% The impact of a PAIS update is then calculated using the DT-IM by analyzing the elements of the OCPN that it modifies (structural impacts), current states affected by the update (operational impacts), and the new value of process diagnostic measures after the update (performance impacts). 
The design search process is inspired by the existing literature on ERP post-implementation change management~\cite{parhizkar2017impact}, from which we draw ideas to model system updates and evaluate their impact.
In this work, we limit the evaluation of the artifact to the feasibility of the proposed approach. 
To this aim, we have built a Web-based impact analysis software artifact and conducted a case study based on a procure-to-pay process loosely modeled following the standard one of the SAP ERP system. 

The remainder is organized as follows. We discuss the related work in \autoref{sec:related}. Then, we present the preliminaries in \autoref{sec:preliminaries}. Next, we introduce the DT-IM in \autoref{sec:dtim} and an approach to impact analysis based on the DT-IM in \autoref{sec:impact-analysis}. 
% to evaluate the current state of business processes with action patterns and produce necessary actions.
Afterward, \autoref{sec:evaluation} introduces the implementation of a web application and a case study using the web application.
Finally, \autoref{sec:conclusion} concludes the paper.

\section{Related Work}\label{sec:related}

Digital twins enabled by increasingly powerful data modeling and analysis capabilities have been envisioned by Gelernter in the 1990s~\cite{gelernter1993mirror} and have found widespread adoption in engineering in the last few years~\cite{eramo2021conceptualizing}. The idea of digital twins of organizations has emerged recently~\cite{caporuscio2020architectural} as a means to address the challenges of information processing in modern digital transformation.

Mendling et al.~\cite{mendling2020building} recently have highlighted an ongoing tension between business process management and digital transformation (DT): processes must be flexible to adapt to the continuously changing requirements of DT, while DT must rely on some process compliance to avoid a continuous disruption of business operations. 

Business processes and the information systems supporting them can evolve to support changing business requirements through configuration and adaptation~\cite{dohring2014configuration,la2011configurable}. The former entails that all the possible evolution options are known a priori and can be captured into configuration tables, whereas adaptation involves ad-hoc modification of the process models or systems. Configuration is obviously more agile, but often it cannot address unexpected situations. 

Business process flexibility can also be achieved through run-time adaptation. Along this direction, van Beest et al.~\cite{van2014automated} have proposed an approach for repairing processes at run-time when they interfere, e.g., when inconsistent writing operations occur. Marrella~\cite{marrella2019automated} have proposed to use automated planning techniques to support process adaptation to changing environments.  

An issue closely related to process and system adaptability is the tracking of their evolution over time, understanding, in particular, the impact of proposed updates. In the context of cross-organizational information systems~\cite{comuzzi2012measures} or multi-tenant cloud systems~\cite{kumara2018runtime}, the evolution of systems and processes can be tracked by the evolution of Service Level Agreements (SLAs), which are updated by changing business requirements.

In the context of ERP systems, Soffer et al.~\cite{soffer2003erp,soffer2005aligning} have proposed an ERP modeling language and a related methodology to align the ERP system capabilities with the enterprise requirements. The modeling language allows to express dependencies between business processes and objects, but it does not allow to express changes of them and their impact. Parhizkar and Comuzzi~\cite{parhizkar2017impact,comuzzi2017methodology} have proposed a methodology and initial tool support to characterize the impact of ERP post-implementation changes inspired by the engineering change management literature. Lin et al.~\cite{lin2021succerp} have proposed a method that supports users during the execution of ERP post-implementation changes but which does not support the evaluation of the impact of such changes.

In summary, methods and tools to holistically support the tracking of information systems updates, understanding, in particular, their impact, remain limited in scope, functionality, and degree of automation. This paper tackles this research gap by proposing to use DTOs to develop tools that can (semi-)automatically support the assessment of PAIS updates.

\section{Background and Preliminaries}\label{sec:preliminaries}

%We introduce here a meta and the digital twin interface model.
This section introduces a conceptual model of PAISs (\autoref{subsec:pais}) and the OCPNs, which we use later to formally model PAISs (\autoref{subsec:ocpn}).
% formal definitions of OCPNs (\autoref{subsec:ocpn}).

\subsection{Process-Aware Information Systems (PAISs)}\label{subsec:pais}
\begin{figure}[!htb]
    \centering
    \includegraphics[width=1\linewidth]{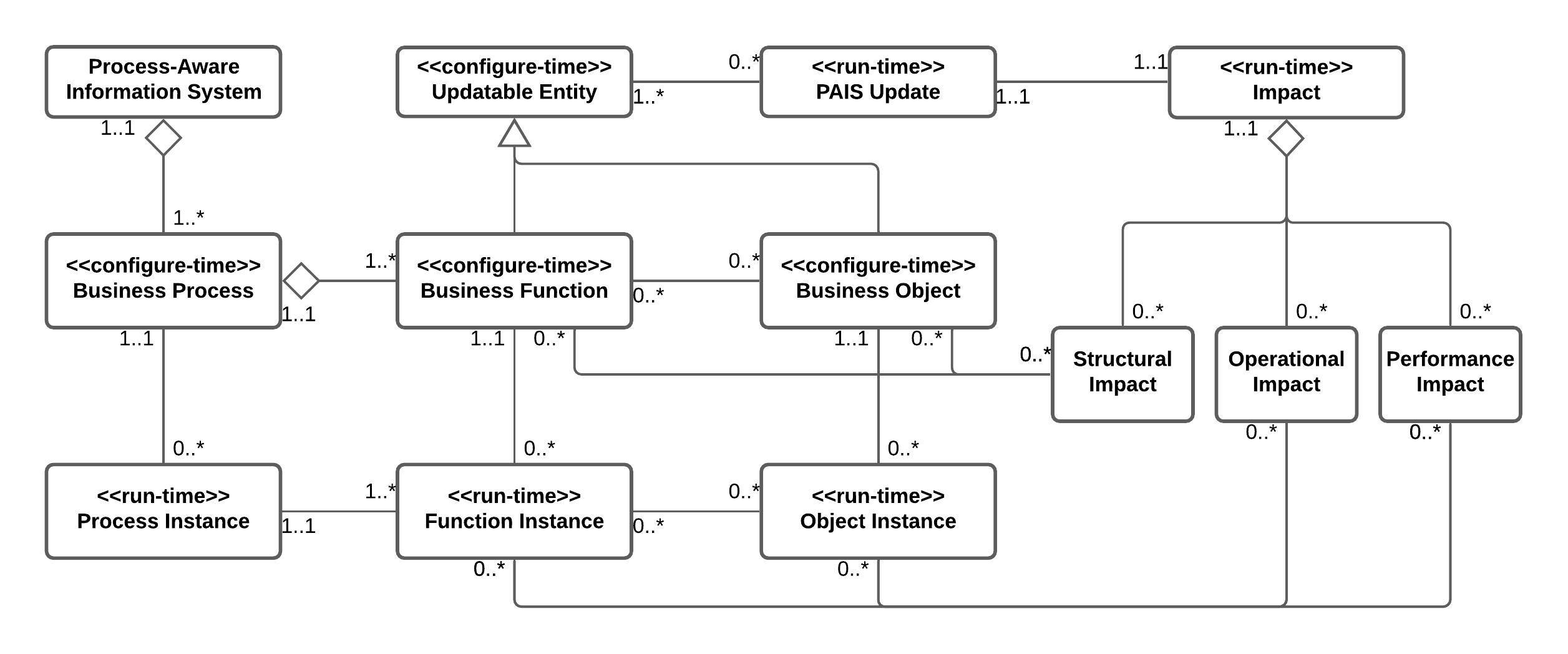}
    \caption{A meta model of PAISs as a \textit{UML 2.0 class diagram}}
    \label{fig:erp-meta-model}
\end{figure}

\autoref{fig:erp-meta-model} introduces a meta-model of PAIS entities, updates, and impacts of PAIS updates considered in this work. 
This is inspired mainly by the work of Parhizkar and Comuzzi~\cite{parhizkar2017impact} in the context of ERP systems. 

% This subsection introduces a meta-model of Process-Aware Information Systems (PAISs) entities, PAIS updates, and impacts of PAIS updates.

% \vspace{-0.5cm}
% \vspace{-0.5cm}

\textbf{PAIS Entities.}\label{sec:erp}
The meta model separates entities concerned with the configure-time from entities related with the run-time. 
At configure-time, a PAIS is constituted by a set of \emph{business processes}, which in turn orchestrate a set of \emph{business functions} that manipulate, using CRUD operations, \emph{business objects}.
% A function object is a special type of object used to define business functions, such as a threshold on the minimum price used to define a function to place an ERP order.
%Note that business functions may be executed when using an ERP system outside of business processes, e.g., a discount may be applied to an existing invoice at any time. 
%However, in this work we assume a process-centric perspective, assuming that all the operations of an organisation are captured by business processes. 
At run-time, a PAIS instantiates instances of processes, functions, and objects.

\textbf{PAIS Updates.}\label{subsec:erp-updates}
A PAIS may be modified by creating, updating, or deleting an existing entity, i.e., an object, function or process. 
In this work, we focus on updates to business functions and objects.
Creating a new entity has no impact on the business operations since it extends the functionality of a system and it only applies to new instances of business processes, functions, or objects. 
Deleting an entity can be seen as a special case of updating it.
Updates to business processes, such as modifications of their control flow, are left out of scope. 

A PAIS update is associated with an updatable entity, i.e., business functions and objects.
First, an \emph{update of a business function} is achieved by changing business rules and functionalities defining it.
% function objects defining it or changing its functionality, which we define here as its set of writing operations.
For instance, a \textit{place order} function in an ERP system can be updated by changing the minimum price to create orders (i.e., a business rule) or changing a set of attributes that the function will update in a business object (i.e., a functionality).
Such updates not only affect the business function, but also instances of the business function.

Second, an \emph{update of a business object} is achieved by adding or removing attributes to it. 
For instance, the \emph{order} business object can be updated by adding \textit{payment terms} to specify the conditions of the payment, e.g., received by the end of the month, within seven days or through monthly installments. 
The update affects both business objects and their instances.

\textbf{Impacts of PAIS Updates.}\label{subsec:impacts}
A PAIS update results in an impact to the system.
% \autoref{fig:impact-types} describes the different types of impacts considered in this work.
Below are the three types of impacts that we consider.
\begin{itemize}
    \item \textit{Structural impacts} concern configure-time entities, i.e., business objects and functions. They include:
    \begin{itemize}
        \item \textit{structural object impact}: the impact on business objects, e.g., the number of impacted business objects, and
        \item \textit{structural function impact}: the impact on business functions, e.g., the number of impacted business functions.
    \end{itemize}
    \item \textit{Operational impacts} concern run-time entities, i.e., the instances of business objects and functions. They include:
    \begin{itemize}
        \item \textit{operational object impact}: the impact on object instances of business objects, e.g., the number of object instances of impacted business objects, and
        \item \textit{operational function impact}: the impact on object instances of business functions, e.g., the number of objects of impacted business functions.
    \end{itemize}
    \item \textit{Performance impacts} concern changes in diagnostic measures of business objects and functions. They include:
    \begin{itemize}
        \item \textit{object performance impact}: the performance impact on business objects, e.g., difference in the avg. service time for a business object before/after updates, and
        \item \textit{function performance impact}: the performance impact on business functions, e.g., difference in the avg. waiting time for a business function before/after updates.
    \end{itemize}
\end{itemize}
\subsection{Object-Centric Petri Nets (OCPNs)}
\label{subsec:ocpn}
In this work, we model a PAIS introduced in \autoref{subsec:pais} with a DT-IM that uses an OCPN as its core formalism.
This subsection introduces OCPNs.
First, a Petri net is a directed bipartite graph of places and transitions. 
A labeled Petri net is a Petri net with the transitions labeled.
% \subsubsection{Object-Centric Event Data}
% \begin{definition}[Universes]\label{def:universes}
% Let $\univ{ei}$ be the universe of event identifiers, 
% $\univ{act}$ the universe of activity names, 
% $\univ{time}$ the universe of timestamps, 
% $\univ{ot}$ the universe of object types, and
% $\univ{oi}$ the universe of object identifiers.
% $\mathit{type} \in \univ{oi} \rightarrow \univ{ot}$ assigns precisely one type to each object identifier.
% $\univ{omap}{=}\{ \mathit{omap} \in \univ{ot}  \not\rightarrow \pow(\univ{oi}) \mid \forall_{\mathit{ot}\in \mathit{dom}(\mathit{omap})}\ \forall_{\mathit{oi}\in \mathit{omap}(\mathit{ot})}\ \mathit{type}(\mathit{oi}){=}\mathit{ot} \}$ is the universe of all object mappings indicating which object identifiers are included per type.
% Let $\univ{val}$ be the universe of attribute values. 
% Let $\univ{vmap}{=}\univ{attr} \nrightarrow \univ{val}$ be the set of all partial functions mapping a subset of attribute names onto the corresponding values. 
% $\univ{event}{=}\univ{ei} \times \univ{act} \times \univ{time} \times \univ{time} \times \univ{omap}$ is the universe of events.
% \end{definition}

% Given $e{=}(ei,act,st,ct,omap) \in \univ{event}$, $\pi_{ei}(e){=}ei$, $\pi_{act}(e){=}act$, $\pi_{st}(e){=}st$, $\pi_{ct}(e){=}ct$, and $\pi_{omap}(e){=}omap$.
% Note that we assume an event involves the lifecycle information, i.e., start and complete timestamp.

% \subsubsection{Object-Centric Petri Nets}
% This section introduces a labeled Petri net and an OCPN.

\begin{definition}[Labeled Petri Net]\label{def:lpn}
Let $\univ{act}$ be the universe of activity names.
A labeled Petri net is a tuple $N {=} (P,T,F,l)$ with $P$ the set of places, $T$ the set of transitions,
$P \cap T {=} \emptyset$, $F\subseteq (P \times T) \cup (T \times P)$ the flow relation, and $l \in T \not\rightarrow \univ{act}$ a labeling function.
% such that $l(t){=}\tau$ if $t \notin dom(l)$.
\end{definition}

A marking $M_{N} \in \bag(P)$ is a multiset of places.
A transition $tr \in T$ is \textit{enabled} in marking $M_{N}$ if its input places contain at least one token.
The enabled transition may \textit{fire} by removing one token from each of the input places and producing one token in each of the output places.

Each place in an OCPN is associated with an object type to represent interactions among different object types. 
The variable arcs represent the consumption/production of a variable amount of tokens in one step.

\begin{definition}[Object-Centric Petri Net]\label{def:oopn}
Let $\univ{ot}$ be the universe of object types.
An \emph{object-centric Petri net} is a tuple $\mathit{ON} {=} (N,\mathit{pt},F_{\mathit{var}})$ where
$N {=} (P,T,F,l)$ is a labeled Petri net, $\mathit{pt} \in P \rightarrow \univ{ot}$ maps places to object types, and $F_{\mathit{var}}\subseteq F$ is the subset of variable arcs.
\end{definition}

\begin{figure}[!htb]
    \centering
    \includegraphics[width=0.8\linewidth]{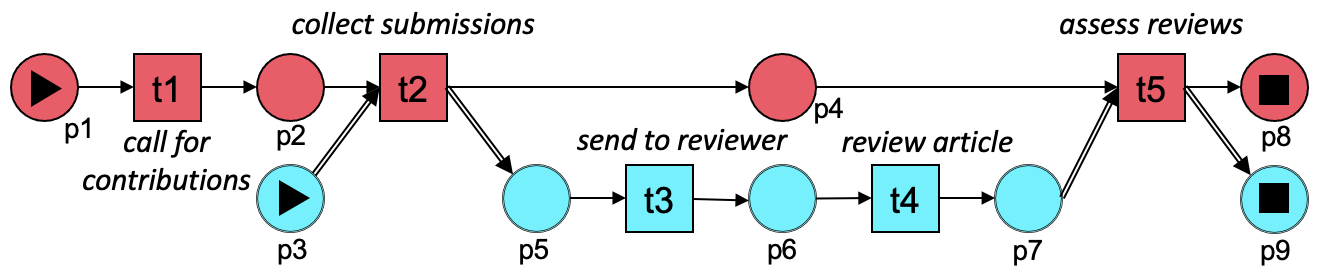}
    \caption{An example of object-centric Petri nets}
    \label{fig:ocpn}
\end{figure}

\autoref{fig:ocpn} shows an OCPN describing a part of the peer review process for an academic conference. There are two types of places associated with two object types, i.e., $conf$ (denoted in red) and $subm$ (cyan). For instance, $pt(p1){=}pt(p2){=}conf$, $pt(p3){=}pt(p5)={subm}$, and $(p3,t2)$, $(t2,p5)$, and $(p7,t5)$ are variable arcs. 

% \vspace{-0.5cm}
% \vspace{-0.75cm}

% , $ON{=}(N,pt,F_{var})$ with $N{=}(P,T,F,l)$ where $P {=} \{ p1, \dots, p9 \}$, $T {=} \{ t1,\dots,t5\}$, $F {=} \{(p1,t1),(t1,p2), \dots \}$, $l(t1){=} \textit{call for contributions}$, etc., $pt(p1) {=} \textit{conference}$, $pt(\\p2){=}\textit{submission}$, etc., and $F_{var} {=} \{(p3,t2), (t2,p5),(p7,t5) \}$.

% A marking of an OCPN represents its state. 

\begin{definition}[Marking]\label{def:oomark}
Let $\univ{oi}$ be the universe of object identifiers.
Let $\mathit{ON}{=}(N,\mathit{pt},F_{\mathit{var}})$ be an object-centric Petri net, where $N{=}(P,T,F,l)$.
% $otyp \in \univ{oi} \to \univ{ot}$ assigns object types to object identifiers.
$Q_{\mathit{ON}}{=}\{(p,\\ \mathit{oi}) \in P \times \univ{oi} \mid type(oi) {=} \mathit{pt}(p)\}$ is the set of possible tokens.
A marking $M$ of $\mathit{ON}$ is a multiset of tokens, i.e., $M \in \bag(Q_{\mathit{ON}})$.
\end{definition}

For instance, marking $M_1{=}[(p4,c1),(p7,s1),(p7,s2)]$ denotes three tokens where place \textit{p4} has one token referring to object \textit{c1} of type \textit{conf} and \textit{p7} has two tokens referring to objects \textit{s1} and \textit{s2} of type \textit{subm}.

% The concept of binding is used to explain the semantics of an OCPN. 
A binding describes the execution of a transition consuming objects from its input places and producing objects for its output places.
A binding $(tr,b)$ is a tuple containing a transition $tr$ and a function $b$ mapping the object types of the input/output places to sets of object identifiers. 
For instance, $(t5,b1)$ describes the execution of transition $t5$ with $b1$ where $b1(\textit{conf}){=}\{c1\}$ and $b1(\textit{subm}){=}\{s1,s2\}$.
%, where \textit{conference} and \textit{submission} are the object types of its surrounding places (i.e., $p4,p7,p8,p9$).

A binding $(tr,b)$ is \textit{enabled} in marking $M$ if all the objects specified by $b$ exist in the input places of $tr$. 
For instance, $(t5,b1)$ is enabled in marking $M_1$ since $c1$, $s1$, and $s2$ exist in its input places, i.e., $p4$ and $p7$.

A new marking $M'$ is reached by executing an enabled binding $(tr,b)$ at marking $M$, denoted by $M \stackrel{(tr,b)}{\longrightarrow} M'$.
For instance, by executing $(t5,b1)$, $c1$ is removed from $p4$ and added to $p8$, while $s1$ and $s2$ are removed from $p7$ and added to $p9$, leading to the new marking $M'{=}[(p8,c1),(p9,s1),(p9,s2)]$.

 Finally, a relation function $rel \in T \to \pow(P)$ maps a transition to a set of places associated with the transition. It is defined recursively: for any $tr \in T$,
% several functions as follows:
% First, $rel \in T \to \pow(P)$ is a relation function and is defined recursively: 
(1) $rel(tr){=}\emptyset$ if $\pre tr {=} \emptyset$ and (2) $rel(tr){=} \pre tr \cup \bigcup_{p \in \pre tr, tr' \in \pre p} rel(tr')$, where $\pre tr$ is a set of input places of $tr$, i.e., $\pre tr{=} \{p \in P \mid (p,tr) \in F \}$.
For instance, $rel(t2){=}\{p1,p2,p3\}$ and $rel(t5){=}\{p1,\dots,p7\}$.
% Second, $M(p)$ is a set of objects in place $p$, i.e., $M(p){=}\{oi \in \univ{oi} \mid (p',oi') \in M \land p{=}p' \land oi{=}oi'\}$.
% For instance, $M_1(p7){=}\{s1,s2\}$.
\section{Modeling PAISs: Digital Twin Interface Model}\label{sec:dtim}
In this work, we use a \textit{digital twin interface model} to model PAIS entities along with PAIS updates introduced in \autoref{subsec:pais}.
% \subsection{Modeling PAISs}
% \label{sec:dto}
\subsection{Modeling PAIS Entities}
A digital twin interface model consists of 1) an OCPN, 2) valves, 3) guards, and 4) operations.
A \emph{valve} is a system configuration, e.g., minimum required quantity to place orders.
A \emph{guard} is a formula composed of attributes, including valves, with relational operators (e.g., $\le$,$\ge$,$=$) and logical operators (e.g., conjunction $\land$, disjunction $\lor$, and negation $\neg$).
$F(X)$ denotes the set of such formulas defined over a set of attributes $X$.
% An \emph{operation} describes the functionality of an activity by specifying which attributes of which object types need to be updated by performing the activity.
An \emph{operation} describes a business operation, e.g., updating the quantity and price of an order.

% A digital twin interface model can be constructed using both event data and domain knowledge. A semi-automated approach to building a digital twin interface model based on event data is presented in~\cite{park-digitaltwin}.
%It first discovers an OCPN describing control-flows of the target business process based on event data and then enriches the net with valves, guards, and operations based on domain knowledge.

\begin{figure}[!htb]
    \centering
    \includegraphics[width=0.9\linewidth]{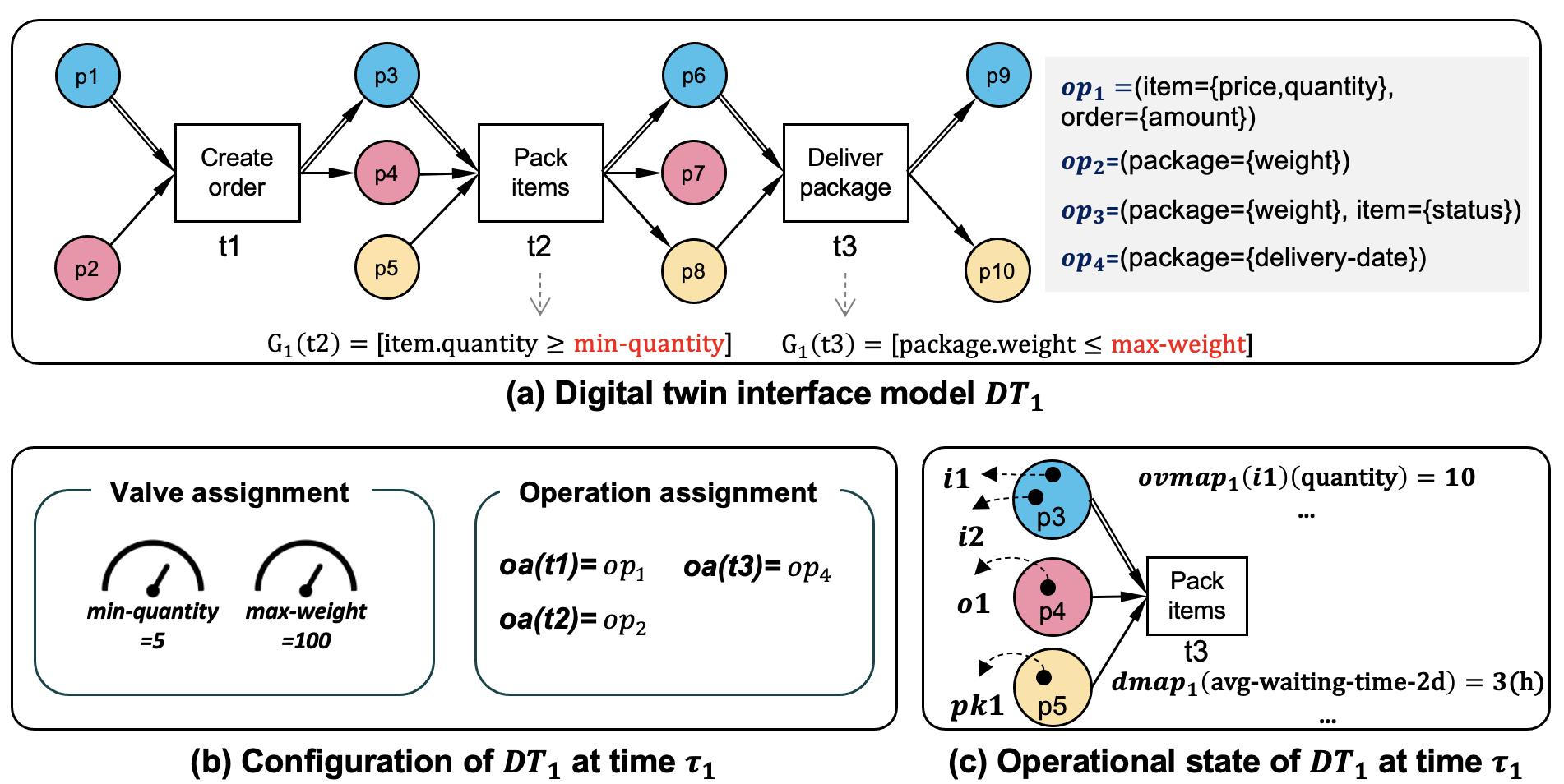}
    \caption{An example of digital twin interface models}
    \label{fig:running-example}
\end{figure}

\begin{definition}[Digital Twin Interface Model (DT-IM)]
Let $\univ{\mi{valve}}$ be the universe of valve names. Let $\univ{\mi{attr}}$ be the universe of attribute names.
A digital twin interface model, denoted as $\mi{DT}$, is a tuple $(\mi{ON},V,A,G,O)$ where
\begin{itemize}
    \item $\mi{ON}{=}(N,pt,F_{\mi{var}})$ is an object-centric Petri net, where $N{=}(P,T,F,l)$,
    \item $V \subseteq \univ{\mi{valve}}$ is a set of valve names,
    \item $A \subseteq \univ{\mi{attr}}$ is a set of attribute names,
    \item $G \in T \to (F(V \cup A) \cup \{\mi{true}\})$ associates transitions with guards, and
    \item $O \subseteq \mi{OT} \nrightarrow \pow(A)$ is a set of operations associating an object type to a set of attributes to be updated, where $\mi{OT}{=}\{pt(p) \mid p \in P\}$. 
    % For any $op \in O$, $op(ot){=}\perp$ if $ot \notin dom(op)$.
\end{itemize}
% $\univ{DT}$ is the set of all possible digital twin interface models.
\end{definition}

The transitions in the OCPN of a DT-IM represent \textit{business functions}, whereas the object types associated with places indicate \emph{business objects}.
Guards and operations represent the business rule(s) and functionality (writing operations) of \emph{business functions}, respectively.
\autoref{fig:running-example}(a) shows an DT-IM, $\mi{DT}_{1}{=}(\mi{ON}_{1},V_1,A_1,\allowbreak G_1,O_1)$, representing a PAIS supporting a simple order management process.
It involves three object types: \textit{item}, \textit{order}, and \textit{package}.
First, an order is created with multiple items. Next, the item is packed and the packaged is delivered to the customer.
Valves are depicted using red italic fonts, and guards are described in squared brackets.
Operations are described in the gray box.
For instance, $op_2 \in O_1$ describes a business operation writing \textit{weight} of \textit{package}, i.e., $op_2(\mi{package}){=}\{\mi{weight}\}$, $op_2(\mi{item}){=}\perp$, and $op_2(\mi{order})=\perp$.

% Note that, in \autoref{fig:dtim}(a), we omit some activity variants and abstract the explanation of them for better representation.
% % For instance, $op_{\textit{vm}}^1,op_{\textit{vm}}^2 \in O_1$ where $op_{\textit{vm}}^1(\textit{material}){=}\{\textit{verified}\}$ and $op_{\textit{vm}}^2(\textit{material}){=}\{\textit{verified},\textit{quality}\}$.
% For instance, $op_1, op_2 \in O_1$ where $op_1(\textit{material}){=}\{\textit{verified}\}$ (i.e., a possible functionality of \textit{verify material}) and $op_2(\textit{material}){=}\{\textit{verified},\textit{quality}\}$ (i.e., another possible functionality of \textit{verify material}).

% Seven valves (depicted as red fonts) are used to define seven guards (described inside squared brackets) and gray dotted lines map transitions to guards.
% For instance $\textit{min-quantity} \in V_1$
% For instance, $G_1(t2){=}[\text{material.net-price}-\text{material.effective-price} \le \textit{max-price-difference} \land \text{material.quantity} \ge \textit{min-quantity}]$.

% For instance, $t2$ is connected to the guard formulating that 1) the difference between the net price and effective price of material should be less than or equal to valve \textit{max-price-difference} and 2) the quantity of the material should be greater than or equal to valve \textit{min-quantity}.

A configuration defines the semantics of a DT-IM by determining the value of valves and the assignment of operations to transitions.

\begin{definition}[Configuration]\label{def:config}
Let $\univ{\mi{val}}$ be the universe of attribute values. 
Let $\mi{DT}{=}(\mi{ON},V,A,G,O)$ be a DT-IM with $\mi{ON}{=}(N,pt,F_{\mi{var}})$ and $N{=}(P,T,F,l)$.
A configuration $\mi{conf}_{\mi{DT}} \in (V \to \univ{\mi{val}}) \times (T \to O)$ is a tuple of a valve assignment $\mi{va}$ and an operation assignment $\mi{oa}$.
% mapping transitions to operations such that, for any $tr {\in} T$, $dom(\mi{oa}(tr)) \subseteq \{ pt(p) \mid p \in \pre tr \}$.
% $\mi{conf} \in V \to \univ{val} \times T \to \pow(\univ{attr})$
% $\mi{conf} \in (V \to \univ{val}) \times (T \to (\univ{ot} \to \pow(\univ{attr})))$ is a tuple of a function mapping valves to values and a function associating transitions with a set of attributes whose values need to be updated.
We denote $\Sigma_{\mi{DT}}{=}(V \to \univ{\mi{val}}) \times (T \to O)$ to be the set of all possible configurations of $\mi{DT}$.
\end{definition}

Given $\mi{conf}_{\mi{DT}}{=}(\mi{va},\mi{oa}) {\in} \Sigma_{\mi{DT}}$, $\pi_{\mi{va}}(\mi{conf}){=}\mi{va}$ and $\pi_{\mi{oa}}(\mi{conf}){=}\mi{oa}$.
Moreover, we denote the configuration of digital twin interface model $\mi{DT}$ at $\tau \in \univ{time}$ as $\mi{conf}_{\mi{DT},\tau}$.
% For instance, \autoref{fig:dtim}(b) describes a configuration of $\mi{DT}_{1}$ at $t \in \univ{time}$, $\mi{conf}_{\mi{DT}_{1},t}$. $\pi_{\mi{va}}(\mi{conf}_{\mi{DT}_{1},t})(\textit{min-quantity}){=}5$, $\pi_{\mi{oa}}(\mi{conf}_{\mi{DT}_{1},t})(t4){=}op_4$, etc.
\autoref{fig:running-example}(b) describes a configuration of $\mi{DT}_{1}$ at $\tau_1 \in \univ{time}$, where $\pi_{\mi{va}}(\mi{conf}_{\mi{DT}_{1},\tau_1})(\textit{min-quantity}){=}5$, $\pi_{\mi{oa}}(\mi{conf}_{\mi{DT}_{1},\tau_1})(\mi{t2}){=}op_2$, etc.

An operational state describes the current status of a business process, i.e., which objects reside in which parts of the process, using the marking of OCPNs.
Moreover, it represents the various diagnostics about the performance of the process, e.g., the average waiting time of activity in the last seven days.
We denote $\Delta_{\mi{DT}}$ to be the set of all possible diagnostics of the digital twin interface model $\mi{DT}$.

\begin{definition}[Operational State of A DT-IM]\label{def:state}
Let $\univ{vmap}{=}\univ{\mi{attr}} \nrightarrow \univ{\mi{val}}$ be the set of all partial functions mapping a subset of attribute names to values. 
Let $\mi{DT}{=}(\mi{ON},V,A,G,O)$ be a DT-IM with $\mathit{\mi{ON}}{=}(N,\mathit{pt},F_{\mathit{var}})$ and $N{=}(P,T,F,l)$. 
An operational state of $\mi{DT}$ is a tuple $\mi{os}_{\mi{DT}}{=}(M,\mi{ovmap},\mi{dmap})$ where
\begin{itemize}
    \item $M \in \bag(Q_{\mathit{ON}})$ is a marking of $\mi{ON}$,
    \item $\mi{ovmap} \in OI \to \univ{vmap}$ is an object value assignment where $OI{=}\{ oi \in \univ{oi} \mid \exists_{p \in P} \; (p,oi) \in M \}$, and
    \item $\mi{dmap} \in \Delta_{\mi{DT}} \nrightarrow \mathbb{R}$ is a diagnostics assignment such that, for any $diag \in \Delta_{\mi{DT}}$, $\mi{dmap}(diag)=\perp$ if $diag \notin dom(\mi{dmap})$.
    % \item $t \in \univ{time}$ is a timestamp
\end{itemize}
% We denote $\Omega_{DT}$ to be the set of all possible states of $DT$.
\end{definition}

Given $\mi{os}_{\mi{DT}}{=}(M,\mi{ovmap},\mi{dmap})$, $\pi_{M}(\mi{os}_{\mi{DT}}){=}M$, $\pi_{\mi{ovmap}}(\mi{os}_{\mi{DT}}){=}\mi{ovmap}$, and $\pi_{\mi{dmap}}(\mi{os}_{\mi{DT}}){=}\mi{dmap}$.
\autoref{fig:running-example}(c) describes an operational state of $\mi{DT}_{1}$ at $\tau_1 \in \univ{time}$, i.e., $\mi{os}_{\mi{DT}_{1}}^1{=}(M_1,\mi{ovmap}_{1},\mi{dmap}_1)$. 
% For instance, $(\mi{i1},\mi{p6}),(\mi{i2},\mi{p6}),(\mi{pk1},\mi{p8}) \in M_1$. 
% The weight of $pk_1$ is $30$, i.e., $\mi{ovmap}_{1}(pk1)(\mi{weight)=30}$.
% The average waiting time for the last 2 days for \textit{delivery package} is $3$ hours, i.e., $\mi{dmap}_1(\textit{avg-waiting-time-2d})=3$.
In the remainder, we denote the operational state of the digital twin interface model $\mi{DT}$ at time $\tau$ as $\mi{os}_{\mi{DT},\tau}$.

We define the semantics of DT-IM by extending the semantics of OCPNs with configurations and operational states.
To this end, we use the notion of digital twin bindings. 

\begin{definition}[Digital Twin Binding]
Let $\mi{DT}{=}(\mi{ON},V,A,G,O)$ be a DT-IM with $\mathit{ON}{=}(N,\mathit{pt},F_{\mathit{var}})$.
A digital twin binding of $\mi{DT}$ is a tuple $((tr,b),w,\tau)$ where $(tr,b)$ is a binding of $\mi{ON}$, $w \in \univ{ot} \to \pow(\univ{\mi{attr}})$ is a write function, and $\tau \in \univ{time}$ is a timestamp.
A digital twin binding $((tr,b),w,\tau)$ is enabled with $\mi{conf}_{\mi{DT},\tau}$ and $\mi{os}_{\mi{DT},\tau}$ if the following conditions are satisfied:
\begin{itemize}
    \item $(tr,b)$ is enabled at $\pi_{M}(\mi{os}_{\mi{DT},\tau})$,
    \item guard $G(tr)$ evaluates to true w.r.t. valve assignment $\pi_{\mi{va}}(\mi{conf}_{\mi{DT},\tau})$ and object value assignment $\pi_{\mi{ovmap}}(\mi{os}_{\mi{DT},\tau})$, and
    \item $w$ corresponds to the assigned operation of $tr$, i.e., $w{=}\pi_{\mi{oa}}(\mi{conf}_{\mi{DT},\tau})(tr)$.
\end{itemize}
\end{definition}

Digital twin binding $((\mi{t2},b),w,\tau_1)$, where $b(\mi{item}){=}\{\mi{i1},\mi{i2}\}$, $b(\mi{order}){=}\{\mi{o1}\}$, $b(\mi{package}){=}\{\mi{pk1}\}$, and $w(\mi{package}){=}\{\textit{delivery-date}\}$, is enabled with the configuration of \autoref{fig:running-example}(b) and the operational state of \autoref{fig:running-example}(c) since $(\mi{t2},b)$ is enabled at $\pi_{M}(\mi{os}_{\mi{DT}_{1},\tau_1})$, $G_1(\mi{t2})$ evaluates to true, and $w$ corresponds to $\pi_{\mi{oa}}(\mi{conf}_{\mi{DT}_{1},\tau_1}\allowbreak)(\mi{t2})$.
% A digital twin binding $(tr,b,w,\tau)$ refers to a binding $(tr,b)$, a write function $w \in \univ{ot} \to \pow(\univ{attr})$ associating an object type to a set of attributes to be updated when firing $tr$, and a timestamp $\tau \in \univ{time}$.
% % Digital twin binding $(tr,b,w') \in T \times (\univ{ot} \to \pow(\univ{oi})) \times (\univ{ot} \times \univ{attr}) \to \univ{val}$
% It is \textit{enabled} if the following conditions are satisfied:
% \begin{itemize}
%     \item $(tr,b)$ is enabled at $\pi_{M}(\mi{os}_{DT,\tau})$,
%     % \item $\mi{conf}_{DT,time}=(\mi{oa},\mi{va}) \in \Sigma_{DT}$,
%     \item guard $G(tr)$ evaluates to true w.r.t. valve assignment $\pi_{\mi{va}}(\mi{conf}_{DT,\tau})$ and $\pi_{\mi{ovmap}}(\mi{os}_{DT,\tau})$, and
%     \item $w$ corresponds to the assigned operation of $tr$, i.e., $w{=}\pi_{\mi{oa}}(\mi{conf}_{DT,\tau})(tr)$.
%     % \item each attribute takes on an admissible value, i.e., for any $\mi{va} \in dom(w)$ (do we need this?), and
% \end{itemize}

\subsection{Modeling PAIS Updates}
Next, we model the PAIS updates introduced in \autoref{subsec:pais} using the notion of \emph{actions} in DT-IMs.
An action updates the configuration of a DT-IM.
First, updating valve assignments of the configuration corresponds to the update of business functions, e.g., updating the value of \textit{min-quantity} changes the business rule of the function \textit{pack items}.
Second, updating operation assignments of the configuration corresponds to both 1) the update of business functions, e.g., updating the operation of \textit{pack items} from $op_2$ to $op_3$ changes the functionality of it (by updating the attribute \textit{status} of items in addition to \textit{weight} of packages), and 2) the update of business objects (by modifying items to have a new attribute \textit{status}).

\begin{figure}[!htb]
    \centering
    \includegraphics[width=0.8\linewidth]{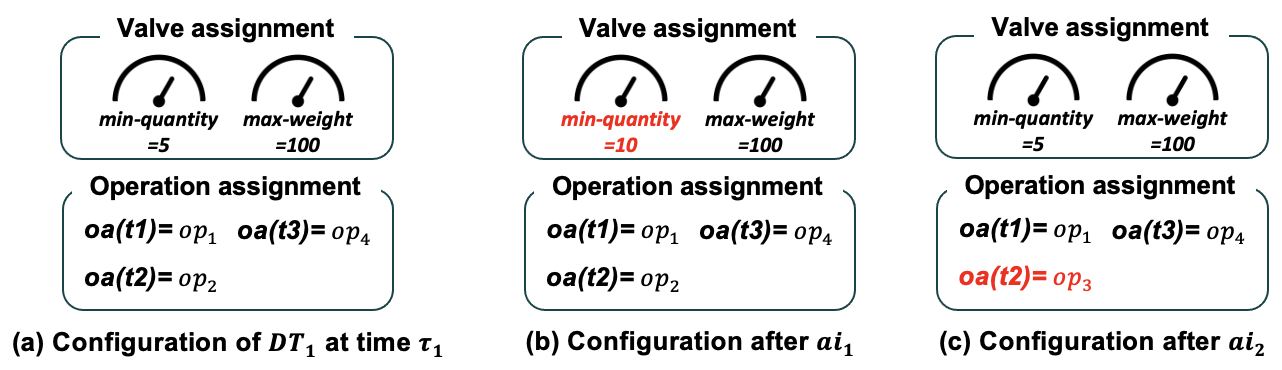}
    \caption{A configuration of $\mi{DT}_{1}$ at $\tau_1$ and new configurations after applying $ai_1$ and $ai_2$}
    \label{fig:action-example}
\end{figure}

% Actions could be either defined manually by domain experts or automatically derived by analyzing event data.

\begin{definition}[Action]\label{def:action}
Let $\mi{DT}$ be a DT-IM. 
An action $\mi{act} \in \Sigma_{\mi{DT}} \to \Sigma_{\mi{DT}}$ updates the configuration.
$A_{\mi{DT}}$ is the set of all possible actions defined over $\mi{DT}$.
\end{definition}

An action instance describes the application of an action. 
An action is applied at a certain start time and, in principle, the configuration change that it entails can remain in place for the foreseeable future until a condition, e.g., on performance metrics, is met or until a specific end time. For simplicity, in this work, we consider only the latter case. 

\begin{definition}[Action Instance]\label{def:action_instance}
Let $\mi{DT}$ be a DT-IM. 
% An action $act \in A_{DT} \to \Sigma_{DT}$ updates the configuration.
An action instance $ai \in A_{\mi{DT}} \times \univ{time} \times \univ{time}$ is a tuple of an action, start timestamp, and end timestamp.
$\mi{AI}_{\mi{DT}}$ is the set of all possible action instances defined over $\mi{DT}$.
\end{definition}

For instance, $ai_1{=}(\mi{act}_{1},\tau_1,\tau_2) \in \mi{AI}_{\mi{DT}_{1}}$ describes the application of $\mi{act}_{1} \in A_{\mi{DT}_{1}}$ to the configuration of $\mi{DT}_{1}$ at $\tau_1$ (\autoref{fig:action-example}(a)) leading to the configuration depicted in \autoref{fig:action-example}(b) until $\tau_2$. 
$ai_2=(\mi{act}_{2},\tau_1,\tau_3) \in \mi{AI}_{\mi{DT}_{1}}$ describes the application of $\mi{act}_{2} \in A_{\mi{DT}_{1}}$ to to the configuration of $\mi{DT}_{1}$ at $\tau_1$ (\autoref{fig:action-example}(a)), producing the configuration shown in \autoref{fig:action-example}(c) until $\tau_3$. 

% For instance, $ai_1=(\mi{act}_{1}, \text{\scriptsize 09:00 20-12-2021}, \text{\scriptsize 09:00 21-12-2021}) \in \mi{AI}_{DT}$ describes the execution of $\mi{act}_{1}$ starting from \text{\scriptsize 09:00 20-12-2021} and finishing at \text{\scriptsize 09:00 21-12-2021}, while $ai_2=(\mi{act}_{2}, \text{\scriptsize 09:00 25-12-2021}, \text{\scriptsize 09:00 31-12-2021}) \in \mi{AI}_{DT}$ represents the execution of $\mi{act}_{2}$ starting from \text{\scriptsize 09:00 25-12-2021} and finishing at \text{\scriptsize 09:00 31-12-2021}.

% \begin{definition}[Changing Valves]\label{def:action}
% Let $DT{=}(ON,V,G)$ be a DT-IM. 
% $chg_{valve} \in \Sigma_{DT} \to \Sigma_{DT}$ updates the configuration.
% % $A_{DT}$ denotes the set of all possible actions defined over $DT$.
% \end{definition}

% \begin{definition}[Changing Write Operations]\label{def:action}
% Let $DT{=}(ON,V,G,W)$ be a DT-IM. 
% $chg_{write} \in (T \to \pow(\univ{attr})) \to (T \to \pow(\univ{attr}))$ updates write operations such that, for any $w \in T \to \pow(\univ{attr})$, $chg_{write}(w)\neq w$.
% \end{definition}

% Action is effective at certain time.
% \begin{definition}[Action Instance]\label{def:action-instance}
% Let $DT$ be a DT-IM. 
% Let $act \in A_{DT}$ be an action.
% An action instance $ai \in A_{DT} \times \univ{time}$ is a tuple of an action and a timestamp.
% \end{definition}

% An action instance leads to an \textit{effective change} in valves and transitions whose values and activity assignments, i.e., depend on the configuration of the DT-IM at the start of the action instance.  

The application of actions results in different \textit{effective changes}, depending on the configuration at the start of the action instance.
% Depending on the configuration at the start of an action instance, its execution results in different changes in the configuration, called \textit{effective changes}.
An effective change of an action instance denotes the valves and transitions whose values and activity assignments are changed due to the action.

% The execution of action instances leads to changes in 1) values of some valves and 2) operations of some transitions.
% Given an action instance, a change function computes the valves and transitions whose values and operations, respectively, are changed due to the action.

\begin{definition}[Effective Change]\label{def:action-instance}
Let $\mi{DT}$ be a DT-IM and $ai{=}(\mi{act},st,ct) \in \mi{AI}_{\mi{DT}}$ an action instance.
An effective change of $ai$ is a tuple of a set of valves and a set of transitions, i.e., $\delta_{ai}{=}(V_c,T_c)$ with $V_c{=}\{v {\in} V \mid \pi_{\mi{va}}(\mi{conf}_{\mi{DT},st})(v) {\neq} \allowbreak \pi_{\mi{va}}(\mi{act}(\mi{conf}_{\mi{DT},st}))(v) \}$ and $T_c{=}\{tr {\in} T \mid \pi_{\mi{oa}}(\mi{conf}_{\mi{DT},st})(tr) {\neq} \pi_{\mi{oa}}(\mi{act}( \mi{conf}_{\mi{DT},st}\allowbreak))(tr) \})$.
% A change function $chg_{DT} \in \mi{AI}_{DT} \to \pow(V) \times \pow(T)$ maps an action instance to a set of valves and a set of transitions, such that, for any $ai \in \mi{AI}_{DT}$, $chg_{DT}(ai)=(V_c,T_c)$ with $V_c=\{v \in V \mid \pi_{\mi{va}}(\mi{conf}_{DT,st})(v) \neq \pi_{\mi{va}}(act(\mi{conf}_{DT,st}))(v) \}$ and $T_c=\{t \in T \mid \pi_{\mi{oa}}(\mi{conf}_{DT,st})(t) \neq \pi_{\mi{oa}}(act(\mi{conf}_{DT,st}))(t) \})$.
\end{definition}

As noted in \autoref{fig:action-example}(b) and \autoref{fig:action-example}(c) with red fonts, the effective change by $ai_1$ is valve \textit{min-quantity}, i.e., $\delta_{ai_1}{=}(\{\textit{min-quantity}\},\emptyset)$, and the effective change by $ai_2$ is the operation assignment of $\mi{t2}$, i.e., $\delta_{ai_2}{=}(\emptyset,\{\mi{t2}\})$.

% Supposing that $\mi{conf}_{\mi{DT}_{1},\text{\scriptsize 09:00 20-12-2021}}$ and $\mi{act}_{1}(\mi{conf}_{\mi{DT}_{1},\text{\scriptsize 09:00 20-12-2021}})$ has only difference in the value of valve \textit{min-quantity}, $\delta_{ai_1}{=}(\{\textit{min-quantity}\},\emptyset)$.
% Moreover, assuming that $\mi{conf}_{\mi{DT}_{1},\text{\scriptsize 09:00 25-12-2021}}$ and $\mi{act}_{2}(\mi{conf}_{\mi{DT}_{1},\text{\scriptsize 09:00 25-12-2021}})$ has only difference in the activity assignment of \textit{verify material}, $\delta_{ai_2}{=}(\emptyset,\{t4\})$.
\section{Impact Analysis}\label{sec:impact-analysis}
% \vspace{-0.75cm}
\begin{figure}[!htb]
    \centering
    \includegraphics[width=1\linewidth]{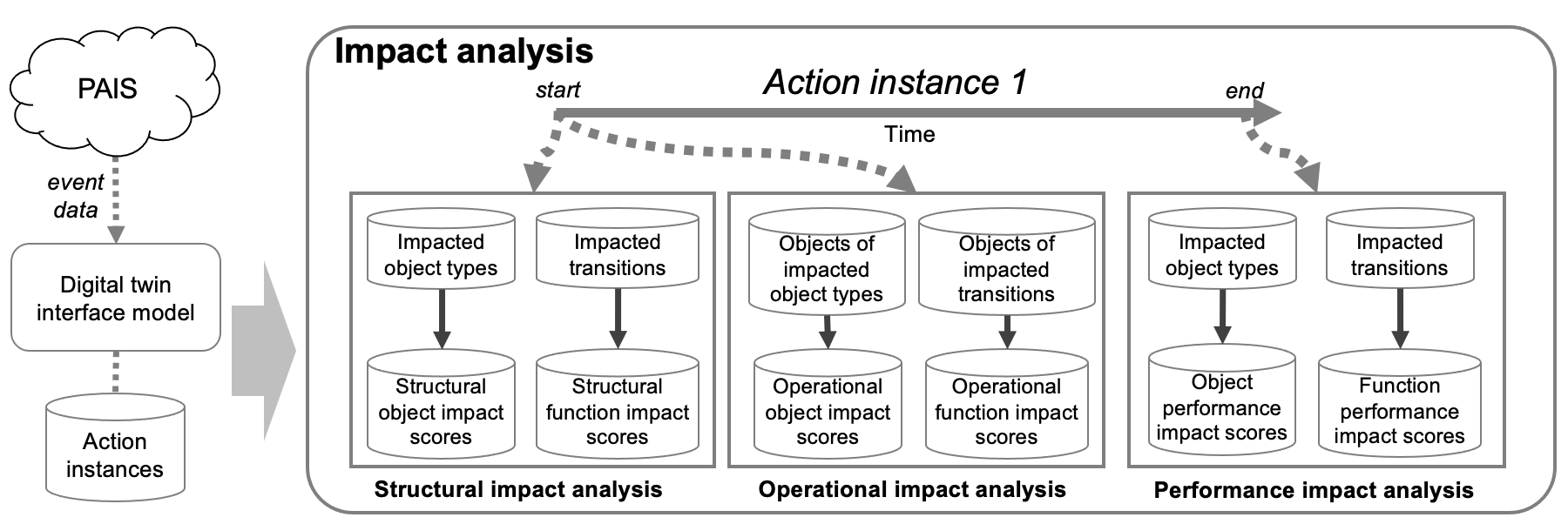}
    \caption{An overview of the proposed impact analysis using digital twin interface models}
    \label{fig:concept}
\end{figure}
% \vspace{-0.5cm}
This section introduces an approach to impact analysis of PAIS updates based on digital twin interface models. 
\autoref{fig:concept} shows an overview of the proposed approach consisting of three components: structural/operational/performance impact analysis.
Using a DT-IM representing a target PAIS, we analyze structural and operational impacts of an action instance, i.e., PAIS updates, at its execution and performance impacts at its completion.
% In each analysis, we identify entities affected by updates and measure impact scores using various metrics.
% The result of the impact analysis is used by a business analyst (domain expert) to decide a correct course of action, e.g., implementing the change, holding the change for a more thorough review, or implementing the change only for new process instances. 
% We leave this type of decision out of the scope of this paper.

\subsection{Structural Impact Analysis}
Structural impact analysis identifies the \emph{structural object/function impacts} of an action instance.
To this end, we identify the object types, i.e., business objects, and transitions, i.e., business functions, affected by an action instance.
% First, we identify the object types of a DT-IM impacted by an action instance.
% Object types are considered as being impacted if they are involved with the effective change in operation assignments.
% If newly assigned operations introduce new attributes or remove existing attributes for an object type, the object type is considered to be affected.
First, an object type is considered to be impacted if the  operations newly assigned by an action instance introduce new attributes to update or remove existing attributes for the object type.
% For instance, by changing the operation of \textit{verify material} to the one newly involving \textit{quality} of \textit{materials}, we affect object type \textit{material} since \textit{quality} needs to be available and even mandatory field of \textit{material}.

\begin{definition}[Impacted Object Types]
Let $\mi{DT}$ be a DT-IM and $ai{=}(\mi{act}, \\ st,ct) \in \mi{AI}_{\mi{DT}}$ an action instance.
$\mi{IOT}_{\mi{DT}}(ai) \subseteq \univ{ot}$ denotes the set of the object types impacted by $ai$, i.e., $\mi{IOT}_{\mi{DT}}(ai){=}\{ ot \in \univ{ot} \mid \delta_{ai}{=}(V_{c},T_{c}) \land tr \in T_{c} \land \pi_{oa}(\mi{conf}_{\mi{DT},st})(tr)(ot) \triangle \pi_{oa}(\mi{act}(\mi{conf}_{\mi{DT},st}))(tr)(ot) \neq \emptyset \} \}$, where $\triangle$ denotes a symmetric difference of sets.
\end{definition}

% \begin{definition}[Identifying Impacted Object Types]
% Let $DT$ be a DT-IM.
% Let $ai=(\mi{act},st,ct) \in \mi{AI}_{DT}$ be an action instance.
% $\mi{IOT}_{DT}(ai) \subseteq \univ{ot}$ denotes the set of the object types impacted by $ai$, i.e., $\mi{IOT}_{DT}(ai)=\{ ot \in \univ{ot} \mid \delta_{ai}{=}(V_{c},T_{c}) \land tr \in T_{c} \land \pi_{oa}(\mi{conf}_{DT,st})(tr)(ot) \triangle \pi_{oa}(\mi{act}(\mi{conf}_{DT,st})(tr))(ot) \neq \emptyset \} \}$, where $\triangle$ denotes a symmetric difference of sets.
% \end{definition}

For instance, $\mi{IOT}_{DT_1}(ai_2){=}\{ \mi{item} \}$ since the effective change of $ai_2$ (i.e., $(\emptyset,\{\mi{t2}\})$) in the operation assignment of $\mi{t2}$ introduces the new attribute \textit{status} of \textit{item}, i.e., $\pi_{oa}(\mi{conf}_{DT_1,\tau_1})(\mi{t2})(item) \triangle \pi_{oa}(\mi{\mi{act}}_{2}(\mi{conf}_{DT_1,\tau_1}))(\mi{t2})(\mi{item}) \neq \emptyset$ (i.e., $\{\mi{status}\} \triangle \emptyset \neq \emptyset$).
% $\{\textit{verified}\} \triangle \{\textit{verified},\textit{quality}\} \\ \neq \emptyset$, where $\{\textit{verified}\}{=}\pi_{oa}(\mi{conf}_{DT_1,\tau_1})(\mi{t2})(material)$ and $\{\textit{verified},\\ \textit{quality}\}{=}\pi_{oa}(\mi{\mi{act}}_{2}((\mi{conf}_{DT_1,\tau_1})(\mi{t2})(material)))$. 
% the changed operation for \textit{verify material} introduces a new attribute of \textit{material}, i.e., $\{\textit{verified}\} \triangle \{\textit{verified},\textit{quality}\} \neq \emptyset$.

% Second, we analyze the structural function impacts of an action instance by identifying transitions of a DT-IM affected by the action instance.
Next, the transitions of a DT-IM are considered to be impacted if they are associated with the effective change in valve assignments and operation assignments.
First, changes in the valve assignment influence transitions by changing the meaning of the guard associated with them, e.g., changing the valve \textit{min-quality} affects the guard of \textit{pack items}.
Second, changes in the operation assignment affect transitions by changing their functionality.

\begin{definition}[Impacted Transitions]
Let $\mi{DT}{=}(ON,V,A,G,O)$ be a DT-IM with $ON{=}(N,pt,F_{var})$.
Let $ai \in \mi{AI}_{\mi{DT}}$ be an action instance.
% Let $\mi{act} \in A_{DT}$ be an action and $conf \in \Sigma_{\mi{DT}}$ a configuration.
$\mi{IT}_{\mi{DT}}(ai) \subseteq T$ denotes the set of the transitions impacted by $ai$, i.e., $\mi{IT}_{\mi{DT}}(ai){=}\{ tr \in T \mid \delta_{ai}{=}(V_{c},T_{c}) \land ((\exists_{v \in V_{c}} \; v \in G(tr)) \lor tr \in T_{c}) \}$.
% $CI_{\mi{act},conf}^{func}=\{ tr \in T \mid (V_{cng},T_{chg})=chg(a,conf) \land (\forall_{v \in V} \; v \in G(tr) \lor tr \in T_{chg}) \}$.
\end{definition}

For instance, $\mi{IT}_{DT_1}(ai_1){=}\{\mi{t2}\}$ since the effective change of $ai_1$ (i.e., $(\{\textit{mi}\\ \textit{n-quantity}\},\emptyset)$) in valve \textit{min-quantity} affects the guard associated to $\mi{t2}$.  
% \textit{min-quantity} determines the guard assigned to $\mi{t2}$, i.e., $G_1(\mi{t2})$. 
Moreover, $\mi{IT}_{DT_1}(ai_2){=}\{\mi{t2}\}$ since the effective change by $ai_2$ (i.e., $\delta_{ai_2}{=}(\emptyset,\{\mi{t2}\})$) change the functionality of $\mi{t2}$.

Once the impacted object types/transitions are identified, various structural object/function impact scores can be measured.
In this work, we focus on basic count-based measures, e.g., how many object types and transitions are impacted. These measures can be absolute or relative, i.e., normalized by the total number of respective entities.

Additional measures can be obtained by applying filtering or prioritizing to count-based measures. For instance, by filtering \emph{financial} objects, such as \emph{invoice}, we can measure the absolute/relative impact on objects that are relevant for the finance department. Prioritizing refers to weighting differently the impact on different types of entities. For instance, a higher weight can be given to the impacts on \textit{verification}-related activities in a process.

\subsection{Operational Impact Analysis}
Operational impact analysis aims to analyze \emph{operational object/function impacts} of an action instance.
To that end, we identify the existing objects of the impacted object types (for the former) and the objects related to the impacted transitions (for the latter) in a DT-IM.
% We analyze the former by first identifying the existing objects of the impacted object types in a DT-IM.
First, to identify the existing objects of the impacted object types, we use markings from the operational states of the DT-IM.
% to compute the existing objects of the impacted object types.

\begin{definition}[Objects of Impacted Object Types]
Let $\mi{DT}{=}(ON,V,A,\\G,O)$ be a DT-IM with $ON{=}(N,pt,F_{var})$.
Let $ai{=}(\mi{act},st,ct) \in \mi{AI}_{\mi{DT}}$ be an action instance and $\mi{IOT}_{\mi{DT}}(ai)$ the impacted object types by $ai$.
$\widehat{\mi{IOT}}_{\mi{DT}}(ai) \subseteq \univ{oi}$ denotes the set of objects of $\mi{IOT}_{\mi{DT}}(ai)$, i.e.,
$\widehat{\mi{IOT}}_{\mi{DT}}(ai){=}\{oi \in \univ{oi} \mid p \in dom(pt) \land pt(p) \in \mi{IOT}_{\mi{DT}}(ai) \land (p,oi) \in \pi_{M}(\mi{os}_{\mi{DT},st}) \}$.
\end{definition}

For instance, $\widehat{\mi{IOT}}_{DT_1}(ai_2)$ is a set of objects associated with all tokens in \textit{item} places, i.e., $\mi{i1},\mi{i2},\dots$ of marking $\pi_{M}(\mi{os}_{DT_1,\tau_1})$.
% , since $\mi{IOT}_{DT_1}(ai_2)=\{material\}$.

Next, we identify objects related to impacted transitions. An object is related to a transition if it may perform the transition in the future. 

\begin{definition}[Objects of Impacted Transitions]
Let $\mi{DT}$ be a DT-IM.
Let $ai{=}(\mi{act},st,ct) \in \mi{AI}_{\mi{DT}}$ be an action instance and $\mi{IT}_{\mi{DT}}(ai)$ the impacted transitions by $ai$.
% Let $\mi{act} \in A_{DT}$ be an action, $conf \in \Sigma_{DT}$ a configuration, and $os=(M,\mi{ovmap},\mi{dmap},t) \in \Omega_{DT}$ an operational state.
$\widehat{\mi{IT}}_{\mi{DT}}(ai) \subseteq \univ{oi}$ denotes the set of objects of $\mi{IT}_{\mi{DT}}(ai)$, i.e.,
$\widehat{\mi{IT}}_{\mi{DT}}(ai){=}\{oi \in \univ{oi} \mid tr \in \mi{IT}_{\mi{DT}}(ai) \land p \in rel(tr) \land (p,oi) \in \pi_{M}(\mi{os}_{\mi{DT},st}) \}$.
\end{definition}

For instance, $\widehat{\mi{IT}}_{DT_1}(ai_1)$ is a set of objects associated with all tokens in $rel(\mi{t2}){=}\{\mi{i1},\mi{i2},\mi{o1},\mi{pk1}\}$ of marking $\pi_{M}(\mi{os}_{DT_1,\tau_1})$.

Based on the objects of impacted object types/transitions, we measure operational object/function impact scores.
As for the structural impact analysis, in this work, we focus on basic count-based measures, e.g., how many objects of impacted object types/transitions are impacted by an update. 

Also in this case, we can apply filtering or prioritizing to define new measures.
For instance, objects can be filtered based on the value of specific attributes or the stage of their lifecycle, e.g., orders higher than a certain amount or from premium customers, or objects for which payments have been cleared. Regarding objects of impacted transitions, we can filter or prioritize objects that lie directly in the queue for the impacted transition, e.g., giving a higher weight to objects currently waiting for the impacted transition.

\subsection{Performance Impact Analysis}
Performance impact analysis aims at analyzing \emph{object/function performance impacts}.
First, to analyze the former, we compare diagnostics related to impacted object types before and after applying an action instance.

We define the object performance impact analysis as follows.
Let $ai{=}(\mi{act},st,ct)$ be an action instance and $\mi{iot} \in \mi{IOT}_{\mi{DT}}(ai)$ an impacted object type. 
Let $\mi{diag}_{\mi{iot}} \in \Delta_{\mi{DT}}$ be a diagnostics relevant to $\mi{iot}$, e.g., the average total service time of $\mi{iot}$.
We measure the performance impact of $ai$ on $\mi{iot}$ w.r.t. $\mi{diag}_{\mi{iot}}$ as follows: $\pi_{\mi{dmap}}(\mi{os}_{\mi{DT},ct})(\mi{diag}_{\mi{iot}}) - \pi_{\mi{dmap}}(\mi{os}_{\mi{DT},st})(\mi{diag}_{\mi{iot}})$
% We measure the performance impact of $ai$ on $\mi{iot}$ w.r.t. $\mi{diag}_{\mi{iot}}$ as follows:
% \begin{align*}
% \pi_{\mi{dmap}}(\mi{os}_{\mi{DT},ct})(\mi{diag}_{\mi{iot}}) - \pi_{\mi{dmap}}(\mi{os}_{\mi{DT},st})(\mi{diag}_{\mi{iot}})
% \end{align*}

Next, we analyze function performance impacts by comparing diagnostics associated with impacted transitions before and after applying an action instance.
Examples of relevant diagnostics are the average service time of the transition, or the average waiting time of the transition.

We formally define the function performance impact analysis as follows.
Let $ai{=}(\mi{act},st,ct)$ be an action instance and $\mi{it} \in \mi{IT}_{\mi{DT}}(ai)$ an impacted transition. 
Let $\mi{diag}_{\mi{it}} \in \Delta_{\mi{DT}}$ be a diagnostics relevant to $\mi{it}$, e.g., the average service time of $\mi{it}$.
We measure the performance impact of $ai$ on $\mi{it}$ w.r.t. $\mi{diag}_{\mi{it}}$ as follows: 
$\pi_{\mi{dmap}}(\mi{os}_{\mi{DT},ct})(\mi{diag}_{\mi{it}}) - \pi_{\mi{dmap}}(\mi{os}_{\mi{DT},st})(\mi{diag}_{\mi{it}})$.
% \begin{align*}
% \pi_{\mi{dmap}}(\mi{os}_{\mi{DT},ct})(\mi{diag}_{\mi{it}}) - \pi_{\mi{dmap}}(\mi{os}_{\mi{DT},st})(\mi{diag}_{\mi{it}}).
% \end{align*}

In this work we consider general purpose diagnostics, such as the average total service time of impacted object types, or the average total waiting time of impacted object types. Other diagnostics can be defined applying the filtering and prioritizing principles introduced earlier, e.g., considering the average total waiting time of objects of a certain type, or giving more weight to the waiting time of certain types of objects. Diagnostics can also be defined on a domain-specific basis, e.g., process-specific KPIs.

\section{Evaluation}\label{sec:evaluation}
This section presents the implementation of the approach presented in this paper and evaluates its feasibility by applying it to a simulated PAIS.

\subsection{Implementation}\label{subsec:implementation}
We have implemented a cloud-based Web service
% \footnote{\label{impacta} Sources, manuals, and a demo video at \url{https://github.com/gyunamister/impacta}.} 
to support the impact analysis with a dedicated user interface.
Sources, manuals, and a demo video are available at \url{https://github.com/gyunamister/impacta}.
The service comprises the following four functional components:
% 1) designing DT-IMs, 2) updating configurations and operational states, 3) defining and executing action instances, and 4) analyzing structural/operational/performance impacts.
% All four functional components are implemented as Python packages. 
% Containerized as a Docker container, the functionality of the proposed approach is structured into a coherent set of microservices that are deployable on any platform.

\textbf{Designing DT-IMs.}
This component supports the design of DT-IMs based on event data and domain knowledge. 
The input is event data of the standard OCEL~\cite{DBLP:conf/adbis/GhahfarokhiPBA21}, valves, guards, and operations in a JSON-based format.
The event data are used to discover an OCPN using the technique introduced in~\cite{DBLP:journals/fuin/AalstB20}, and valves, guards, and operations enhance the discovered OCPN, completing the design of a DT-IM.

% Below is an example of the guards in JSON-based format:
% \begin{minted}
% [
% frame=lines,
% framesep=2mm,
% baselinestretch=1.2,
% fontsize=\scriptsize
% ]
% {json}
% "guards": [
%     {
%         "transition": "place_order",
%         "guard": "[order.price >= po_price]"
%     },
%     ...
% ]
% \end{minted}
% Below is an example of the valves in JSON-based format:
% \begin{minted}
% [
% frame=lines,
% framesep=2mm,
% baselinestretch=1.2,
% fontsize=\scriptsize
% ]
% {json}
% "valves": {
%     "po_price": {
%         "r_min": 0,
%         "r_max": 100,
%         "default": 5
%     },
%     ...
% }

% \end{minted}
\textbf{Updating Configurations and Operational States.}
This component updates the configuration and operational state of a DT-IM in sync with the updates in a target PAIS.
To this end, it is connected with the PAIS, specifying:  1) the source of the current setting of the PAIS and 2) the source of the streaming event data from the PAIS.
Using the current setting, the configuration of the DT-IM is updated. 
Then, the operational state is updated by replaying the streaming event data using the token-based replay technique described in~\cite{token-based}.
% An approach to compute operational states based on streaming event data is discussed in~\cite{park-digitaltwin}. The marking is computed by projecting event data to the OCPN while the diagnostics are computed by token-based replay~\cite{token-based}.

\textbf{Defining and Executing Actions.}
The goal of this component is to 1) define actions based on the available valves and operations and 2) instantiate them as action instances by specifying start and completion times.
To this end, the service provides visual information to support the definition of actions and action instances.
Once executed, an action instance changes the configuration setting of the system and, accordingly, the configuration of the DT-IM.

% \vspace{-0.5cm}
\begin{figure}[!htb]
    \centering
    \includegraphics[width=1\textwidth]{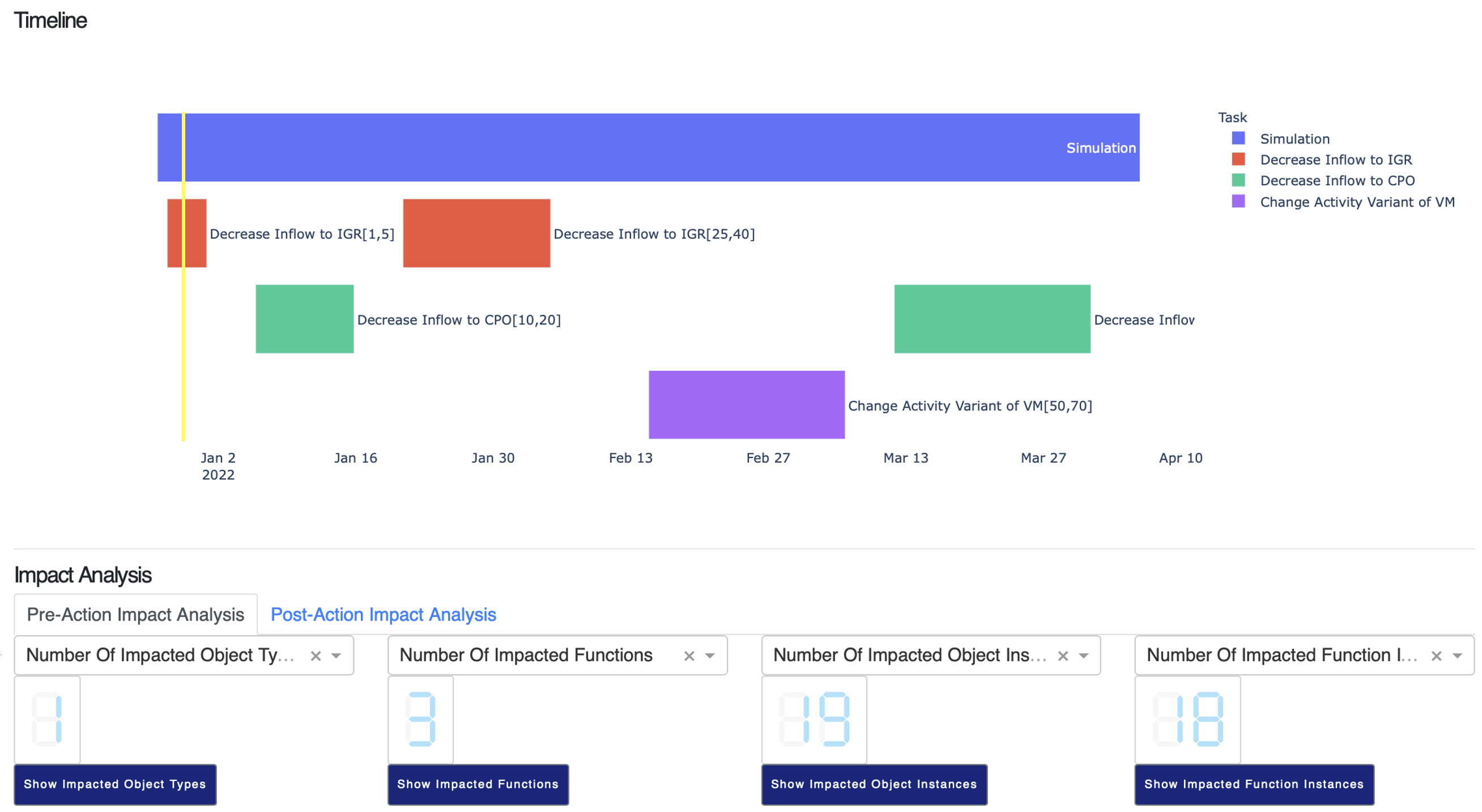}
    \caption{Screenshot of the implementation}
    \label{fig:implementation}
\end{figure}
% \vspace{-0.5cm}

\textbf{Analyzing Structural/Operational/Performance Impacts.}
This component evaluates the impact of action instances.
\autoref{fig:implementation} shows a screenshot of the Web service's interface.
The timeline in the upper part shows the current status, including the current timestamp (yellow vertical line) and the overview of the action instances.
Specifically, five action instances (horizontal bars) of three different actions (distinguished by colors) are scheduled to be executed.
For instance, the first action instance named \textit{Decrease Inflow to IGR} is effective from 26-12-2021 to 31-12-2021.
Note that we compute structural/operational impacts at the start of action instances and performance impacts at the end of them.

% the former shown in \textit{Pre-Action Impact Analysis} tab and the latter shown in \textit{Post-Action Impact Analysis} tab.

\subsection{Case Study}\label{sec:case-study}
Using the implementation, we have conducted a case study on an artificial PAIS that supports a procure-to-pay process.
The system is developed to reflect a real-life SAP ERP system supporting a procure-to-pay processes by using the same business objects, functions, and rules found in SAP.
Using the artificial PAIS, we simulate the procure-to-pay process with 24 resources with different capacities and performance.
Purchase orders are created by following the exponential distribution; the business hours are set as 9-17 from Monday to Friday;
work assignments are scheduled using the \textit{First-in-First-out} rule.
% , which is loosely modelled on the corresponding process in SAP.
% The process in the target PAIS is simulated using the business rules and the operations specified in \autoref{fig:dtim}. 
% We use the configuration described in \autoref{fig:dtim}(b) as a default configuration.

% The process runs using the default configuration unless actions are activated.
% The arrival time of new instances follows the exponential distribution, and the business hours are set as 9-17 from Monday to Friday;
% 24 resources with different capacity and performance serve the process.
% Work assignments are scheduled using the \textit{First-in-First-out} rule.

\begin{figure}[!htb]
    \centering
    \includegraphics[width=1\linewidth]{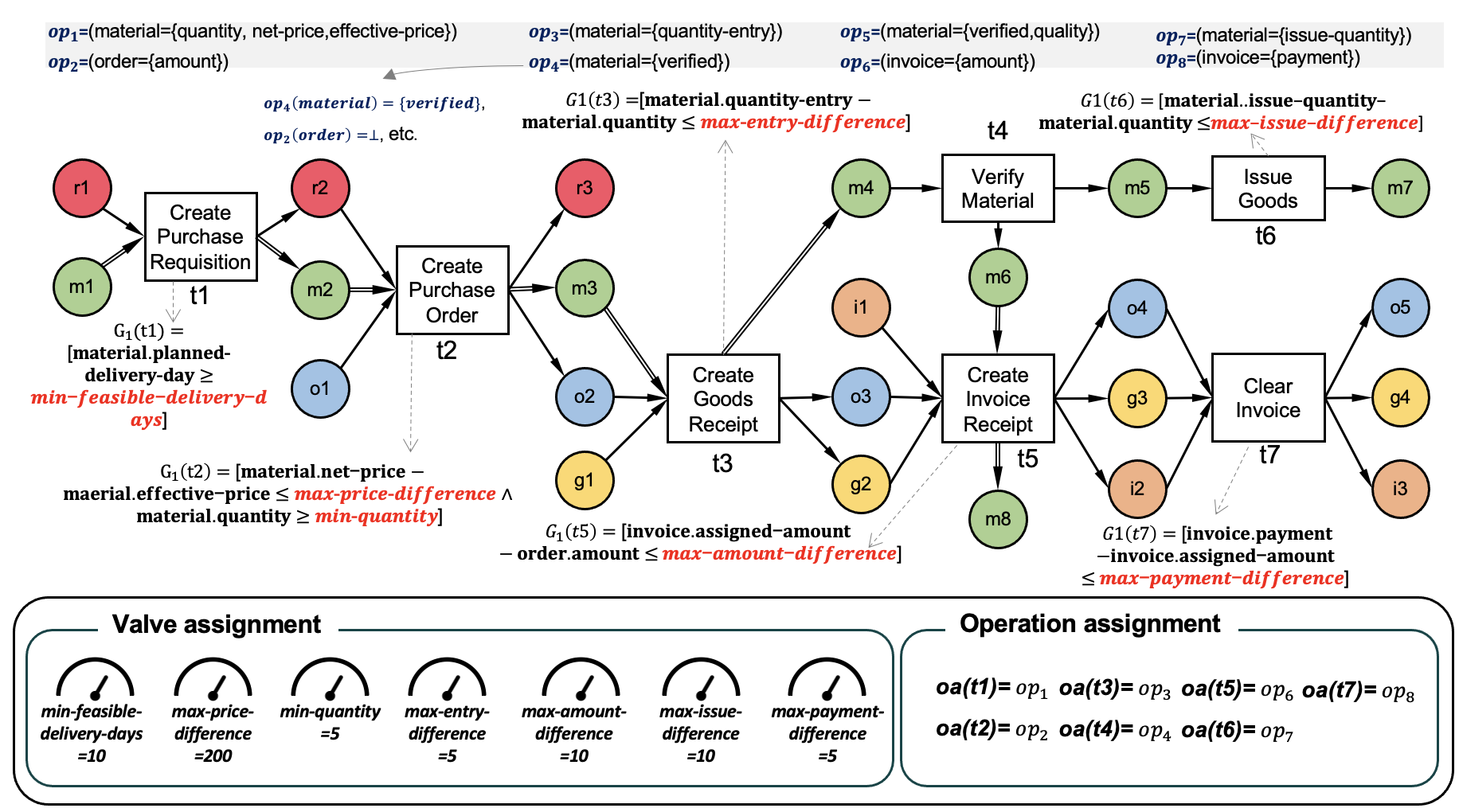}
    \caption{A digital twin interface model of the PAIS supporting a procure-to-payment process and its configuration}
    \label{fig:dtim}
\end{figure}

\autoref{fig:dtim}(a) shows the DT-IM representing the PAIS ($DT_{\mi{p2p}}$).
The process involves five object types.
First, a purchase requisition is created with multiple materials. 
Next, a purchase order is created based on the purchase requisition and material.
A goods receipt is produced after receiving the materials of the purchase order.
Afterward, the material is verified and issued for various purposes, and concurrently the invoice for the purchase order is received.
Finally, the invoice is cleared.
\autoref{fig:dtim}(b) shows the default configuration of the system.

% \vspace{-0.7cm}
% \vspace{-0.6cm}

Using the DT-IM and configuration, we define actions, $A_1$ and $A_2$, as follows:
\begin{itemize}
    \item $A_1$ increases valve \textit{min-quantity} to 10 to reduce the inflow to \textit{create purchase order}, i.e., for any $conf_{DT_{\mi{p2p}}} \in \Sigma_{DT_{\mi{p2p}}}$, $A_1(conf_{DT_{\mi{p2p}}}){=}conf_{DT_{\mi{p2p}}}^{A_1}$ such that $\pi_{va}(conf_{DT_{\mi{p2p}}}^{A_1})(\textit{min-quantity}){=}10$, and
    \item $A_2$ changes the operation of \textit{verify materials} to $op_5$ to additionally update \textit{quality} of materials, i.e., for any $conf_{DT_{\mi{p2p}}} \in \Sigma_{DT_{\mi{p2p}}}$, $A_2(conf_{DT_{\mi{p2p}}}){=}conf_{DT_{\mi{p2p}}}^{A_2}$ such that $\pi_{oa}(conf_{DT_{\mi{p2p}}}^{A_2})(t4){=}op_5$, where $op_5(\textit{material}){=}\{\textit{verified},\textit{quality}\}$.
\end{itemize}

Using the actions, we define the following action instances: $AI_1{=}(A_1,1,5)$ and $AI_2{=}(A_2,10,15)$. 
Note that, for the ease of the simulation, we abstract the timestamp to time steps each of which has the scale of 24 hours, starting from \text{\scriptsize 09:00 20-12-2021}.
For instance, $AI_1$ is effective from \text{\scriptsize 09:00 21-12-2021} to \text{\scriptsize 09:00 25-12-2021}.

% We analyze the structural/operational/performance impact of $ai_1$ and $ai_2$ using the following metrics:
% \begin{itemize}
%     \item Structural object impact: \textit{Total number of impacted business objects}
%     \item Structural function impact: \textit{Total number of impacted business functions}
%     \item Operational object impact: \textit{Total number of object instances of the impacted business objects}
%     \item Operational function impact: \textit{Total number of object instances of impacted business functions}
%     \item Object instance performance impact: \textit{Difference in total service time}
%     \item Object performance impact: \textit{Difference in avg. total service time}
%     \item Function performance impact: \textit{Difference in avg. waiting time}
% \end{itemize}

% \vspace{-0.75cm}
\begin{table}[!htb]
\caption{Results of the impact analysis on $AI_1$ and $AI_2$}
\centering
\begin{adjustbox}{width=\textwidth}
\label{tab:ai-impact}
\begin{tabular}{|c|c|c|c|}
\hline
\textbf{Impact}                                   & \textbf{Metric}                                                   & \textbf{AI 1} & \textbf{AI 2} \\ \hline
Structural object impact                          & Total number of impacted business objects                         & 0             & 1             \\ \hline
Structural function impact                        & Total number of impacted business functions                       & 1             & 1             \\ \hline
Operational object impact                         & Total number of object instances of the impacted business objects & 0             & 254           \\ \hline
Operational function impact                       & Total number of object instances of impacted business functions   & 51            & 153           \\ \hline
Object performance impact                         & Difference in avg. total service time for purchase orders         & -4m           & -              \\ \hline
\multicolumn{1}{|l|}{Object performance impact}   & Difference in avg. total service time for materials               & -             & 1.4h         \\ \hline
\multicolumn{1}{|l|}{Function performance impact} & Difference in avg. sojourn time of create purchase order          & -9m           & -             \\ \hline
\multicolumn{1}{|l|}{Function performance impact} & Difference in total number of purchase orders                     & -13           & -             \\ \hline
Function performance impact                       & Difference in avg. sojourn time of verify material                & -             & 1.6h         \\ \hline
\end{tabular}
\end{adjustbox}
\end{table}
% \vspace{-0.5cm}

\autoref{tab:ai-impact} shows the result of the impact analysis on $AI_1$ and $AI_2$. 
The action instance $AI_1$ affects one transition, i.e., \textit{create purchase order} and 51 running objects related to it.
As a result of the action, the average total service time of purchase orders has been improved by 4 minutes, while the average sojourn time of \textit{create purchase order} has been reduced by 9 minutes.
This is due to the decrease in the queue for the activity, resulting from the new business rule setting the higher minimum quantity for creating purchase orders.
Moreover, the number of purchases has been reduced by 13 during the execution of the action.

The action instance $AI_2$ affects one object type and one transition, i.e., \textit{material} and \textit{verify material}.
Besides, 254 running objects of \textit{material} and 153 objects of \textit{verify material} have been affected by the action.
As a result of the action, the average total service time of purchase orders has increased of 1.4 hours, while the average sojourn time of \textit{verify material} has  increased of 1.6 hours.
% \usepackage{multirow}

% \begin{table}[]
% \centering
% \begin{adjustbox}{width=\textwidth}
% \label{tab:ai2-impact}
% \begin{tabular}{|c|c|c|c|}
% \hline
% \textbf{Type}                & \textbf{Impact}             & \textbf{Metric}                                                   & \textbf{AI 2} \\ \hline
% \multirow{2}{*}{Structural}  & Structural object impact    & Total number of impacted business objects                         & 1             \\ \cline{2-4} 
%                              & Structural function impact  & Total number of impacted business functions                       & 1             \\ \hline
% \multirow{2}{*}{Operational} & Operational object impact   & Total number of object instances of the impacted business objects & 254           \\ \cline{2-4} 
%                              & Operational function impact & Total number of object instances of impacted business functions   & 153           \\ \hline
% \multirow{2}{*}{Performance} & Object performance impact   & Difference in avg. total service time for materials               & -1.4h         \\ \cline{2-4} 
%                              & Function performance impact & Difference in avg. sojourn time                                   & -1.6h         \\ \hline
% \end{tabular}
% \end{adjustbox}
% \end{table}

\section{Conclusions}\label{sec:conclusion}
In this paper, we proposed an approach to impact analysis of PAIS updates based on a DT-IM.
PAIS updates are modeled as updates of the configuration in a DT-IM. 
Next, we identify PAIS entities impacted by PAIS updates and measure the structural/operational/performance impacts based on such entities.
We have implemented the approach as a Web application and discussed a case study on a standard Procure-to-Pay process.

The proposed approach has several limitations.
First, the identification of objects related to impacted business object types and transitions is limited to existing objects in the process and the future objects entering the process are not considered.
Second, we identify objects related to impacted transitions, including all objects that potentially execute the transition.
However, some objects may bypass the transition, e.g., a patient expected to perform surgery may die before it, or a doctor may decide for an emergency treatment at the last moment.
% Second, the identification of run-time entities are computed by including all object instances 
% the conditional execution of the process control flow is not taken into account when considering the impact analysis. For instance, the object instances of impacted business functions are computed, including all object instances that potentially execute a business function.
% However, some object instances may bypass the business function, e.g., a patient expected to perform surgery may die before it, or a doctor may decide for an emergency treatment at the last moment. 
Finally, the performance impact analysis does not isolate the objects subject to action instances to evaluate the performance impact, instead indirectly evaluating changes in diagnostics over all existing objects in the process.

Besides addressing the above limitations, as future work, we plan to extend the approach to predict the performance impact to provide timely and accurate information before the execution of any update.
%We also plan to improve the computation of object instances of impacted business functions such that it only counts that certainly perform the business function.
Another direction of future work is to improve the performance impact analysis such that it completely isolates the instances affected by changes to provide more realistic performance impact measures. Finally, we also plan to evaluate the proposed approach's ease of use and usefulness with business analysts in real-world situations. 

%\vspace{-0.3cm}
%\section*{Acknowledgment}
%\vspace{-0.1cm}
%This work is supported by the Alexander von Humboldt (AvH) Stiftung for funding this research and the 0000 Project Fund (Project n. 1.220047.01) of UNIST (Ulsan National Institute of Science \& Technology).

\bibliographystyle{splncs04}
\bibliography{mybib}

\begin{thebibliography}{10}
\providecommand{\url}[1]{\texttt{#1}}
\providecommand{\urlprefix}{URL }
\providecommand{\doi}[1]{https://doi.org/#1}

\bibitem{DBLP:journals/fuin/AalstB20}
van~der Aalst, W.M.P., Berti, A.: Discovering object-centric {Petri} nets.
  Fundam. Informaticae  \textbf{175}(1-4),  1--40 (2020)

\bibitem{van2014automated}
van Beest, N.R., Kaldeli, E., Bulanov, P., Wortmann, J.C., Lazovik, A.:
  Automated runtime repair of business processes. Information Systems
  \textbf{39},  45--79 (2014)

\bibitem{token-based}
Berti, A., van~der Aalst, W.M.P.: A novel token-based replay technique to speed
  up conformance checking and process enhancement. Trans. Petri Nets Other
  Model. Concurr.  \textbf{15},  1--26 (2021)

\bibitem{caporuscio2020architectural}
Caporuscio, M., Edrisi, F., Hallberg, M., Johannesson, A., Kopf, C.,
  Perez-Palacin, D.: Architectural concerns for digital twin of the
  organization. In: European Conference on Software Architecture. pp. 265--280.
  Springer (2020)

\bibitem{comuzzi2017methodology}
Comuzzi, M., Parhizkar, M.: A methodology for enterprise systems
  post-implementation change management. Industrial Management \& Data Systems
  (2017)

\bibitem{comuzzi2012measures}
Comuzzi, M., Vonk, J., Grefen, P.: Measures and mechanisms for process
  monitoring in evolving business networks. Data \& knowledge engineering
  \textbf{71}(1),  1--28 (2012)

\bibitem{dohring2014configuration}
D{\"o}hring, M., Reijers, H.A., Smirnov, S.: Configuration vs. adaptation for
  business process variant maintenance: an empirical study. Information Systems
   \textbf{39},  108--133 (2014)

\bibitem{dumas2005process}
Dumas, M., van~der Aalst, W.M.P., Hofstede, A.H.T.: Process-aware information
  systems: bridging people and software through process technology. John Wiley
  \& Sons (2005)

\bibitem{eramo2021conceptualizing}
Eramo, R., Bordeleau, F., Combemale, B., van Den~Brand, M., Wimmer, M.,
  Wortmann, A.: Conceptualizing digital twins. IEEE Software  (2021)

\bibitem{gelernter1993mirror}
Gelernter, D.: Mirror worlds: Or the day software puts the universe in a
  shoebox... How it will happen and what it will mean. Oxford University Press
  (1993)

\bibitem{DBLP:conf/adbis/GhahfarokhiPBA21}
Ghahfarokhi, A.F., Park, G., Berti, A., van~der Aalst, W.M.P.: {OCEL:} {A}
  standard for object-centric event logs. In: New Trends in Database and
  Information Systems. vol.~1450, pp. 169--175 (2021)

\bibitem{kumara2018runtime}
Kumara, I., Han, J., Colman, A., van~den Heuvel, W.J., Tamburri, D.A.: Runtime
  evolution of multi-tenant service networks. In: European Conference on
  Service-Oriented and Cloud Computing. pp. 33--48 (2018)

\bibitem{la2011configurable}
La~Rosa, M., Dumas, M., Ter~Hofstede, A.H., Mendling, J.: Configurable
  multi-perspective business process models. Information Systems
  \textbf{36}(2),  313--340 (2011)

\bibitem{lin2021succerp}
Lin, Y.Y., Nagai, Y., Chiang, T.H., Chiang, H.K.: Succerp: The design science
  based integration of ecs and erp in post-implementation stage. International
  Journal of Engineering Business Management  \textbf{13},  18479790211008812
  (2021)

\bibitem{marrella2019automated}
Marrella, A.: Automated planning for business process management. Journal on
  data semantics  \textbf{8}(2),  79--98 (2019)

\bibitem{mendling2020building}
Mendling, J., Pentland, B.T., Recker, J.: Building a complementary agenda for
  business process management and digital innovation (2020)

\bibitem{parhizkar2017impact}
Parhizkar, M., Comuzzi, M.: Impact analysis of erp post-implementation
  modifications: Design, tool support and evaluation. Computers in Industry
  \textbf{84},  25--38 (2017)

\bibitem{park-digitaltwin}
Park, G., van~der Aalst, W.M.P.: Realizing a digital twin of an organization
  using action-oriented process mining. In: ICPM 2021. pp. 104--111 (2021)

\bibitem{soffer2003erp}
Soffer, P., Golany, B., Dori, D.: Erp modeling: a comprehensive approach.
  Information systems  \textbf{28}(6),  673--690 (2003)

\bibitem{soffer2005aligning}
Soffer, P., Golany, B., Dori, D.: Aligning an erp system with enterprise
  requirements: An object-process based approach. Computers in industry
  \textbf{56}(6),  639--662 (2005)

\bibitem{wieringa2014design}
Wieringa, R.J.: Design science methodology for information systems and software
  engineering. Springer (2014)

\end{thebibliography}

\end{document}